\pdfoutput=1
\documentclass[11pt]{article}
\usepackage[final]{acl}
\usepackage{times}
\usepackage{latexsym}
\usepackage[T1]{fontenc}
\usepackage[utf8]{inputenc}
\usepackage{microtype}
\usepackage{inconsolata}
\usepackage{graphicx}
\usepackage{algorithm}
\usepackage{algpseudocode} 
\usepackage{amsmath}
\usepackage{amssymb}
\usepackage{amsthm}
\theoremstyle{plain}
\newtheorem{theorem}{Theorem}[section]
\newtheorem{lemma}[theorem]{Lemma}
\DeclareMathOperator*{\argmin}{arg\,min}
\DeclareMathOperator*{\argmax}{arg\,max}
\DeclareMathOperator*{\E}{\mathbf{E}}
\DeclareMathOperator*{\D}{\mathbf{D}_{KL}}

\newcommand{\pr}[1]{\mathbf{Pr}\left\{#1\right\}}
\newcommand{\N}{\mathbb{N}}
\newcommand{\simple}{\textit{simple} }
\newcommand{\sequence}{\textit{sequence} }
\newcommand{\batch}{\textit{batch} }
\newcommand{\tree}{\textit{tree} }
\newcommand{\simplex}{\textit{simple}}
\newcommand{\sequencex}{\textit{sequence}}

\newcommand{\idx}{\mathbf{j}}
\newcommand{\ent}{\mathbf{H}}
\newcommand{\entb}[1]{\ent\left[#1\right]}
\newcommand{\xd}{\tilde{x}}

\newcommand{\cat}{||}
\usepackage{pgfplots}
\pgfplotsset{width=10cm,compat=1.18}
\usetikzlibrary{arrows.meta, decorations.pathreplacing}
\usetikzlibrary{backgrounds}
\usetikzlibrary{positioning}
\usepackage{subcaption}
\usepackage[shortcuts]{glossaries}
\newacronym{llm}{LLM}{large language model}
\newacronym{gsd}{GSD}{greedy speculative decoding}
\newacronym{ersd}{ERSD}{exponential race speculative decoding}
\definecolor{rejColor}{HTML}{FF4444}
\definecolor{accColor}{HTML}{85C247}
\definecolor{lightBlueColor}{HTML}{ADE4FF}
\title{Speculative Decoding via Exponential Races}
\author{Szymon Kobus \\
  Imperial College London \\
  \texttt{szymon.kobus17@imperial.ac.uk} \\\And
  Deniz G\"und\"uz \\
  Imperial College London}
\begin{document}
\maketitle
\begin{abstract}
Speculative decoding accelerates large language model inference using a smaller draft model. In this paper, we establish a surprising connection between speculative decoding and channel simulation, which aims at simulating a noisy channel using as few bits as possible.  This connection allows us to provide an information-theoretic analysis of the speed up that can be achieved by speculative decoding. Leveraging this link, we derive an explicit relation between generation speed-up and the number of tokens $k$ generated by the draft model for large $k$, which serves as an upper bound for all $k$. We also propose a novel speculative decoding method via exponential races \acs{ersd} that matches state-of-the-art performance. 
\end{abstract}
\begin{figure*}[t]
    \centering
    \begin{minipage}{0.36\textwidth} 
        \centering
        \subcaptionbox{Autoregressive \label{fig:autoregressive}}{
        \begin{tikzpicture}[
          node distance=0.8cm and 1cm,
          dot/.style={circle,inner sep=2pt, label={center:}}
        ]
          \node (root) [dot, fill=black, opacity=0.3] at (0, 0) {};
          \node (node1) [dot, fill=lightBlueColor] at (1.2, 0) {};
          \node[inner sep=0pt, minimum size=0pt] at (1.7, -0.935) {};
          \node[inner sep=0pt, minimum size=0pt] at (-0.5, 0) {};
        
          \draw[-Latex, black] (root) to node [below, black] {start} (node1);
          \vspace{1cm}
        \end{tikzpicture}
        }
        \subcaptionbox{\simple ($k=1$)\label{fig:simple}}{
        \begin{tikzpicture}[
          node distance=0.8cm and 1cm,
          dot/.style={circle,inner sep=2pt, label={center:}}
        ]
          \node (root) [dot, fill=black, opacity=0.3] at (0,0) {};
          \node (node1) [dot, fill=black, label=above:{$\xd_{(1)}$}] at (1.2, 0) {};
          \node (node2) [dot, fill=lightBlueColor] at (1.2, -0.8) {};
          \node (node3) [dot, fill=lightBlueColor] at (2.4, 0) {};
        
          \draw[-Latex, black] (root) to node [below, black] {start} (node1);
        
          \draw[-Latex, accColor] (node1) to node [pos=0.5, above, accColor] {acc} (node3);
          \draw[-Latex, rejColor] (node1) to node [pos=0.5, right, rejColor] {rej} (node2);
        \end{tikzpicture}
        }
        \bigskip 
        \subcaptionbox{\sequence ($k=4$) \label{fig:sequence}}{
            \begin{tikzpicture}[
              node distance=0.8cm and 1cm,
              dot/.style={circle,inner sep=2pt, label={center:}}
            ]
              
              \node (root) [dot, fill=black, opacity=0.3] at (0,0) {};
              \node (n1)   [dot, fill=black, right=of root] {};
              \node (n2)   [dot, fill=black, right=of n1] {};
              \node (n3)   [dot, fill=black, right=of n2] {};
              \node (n4)   [dot, fill=black, right=of n3] {};
              \node (n5)   [dot, fill=lightBlueColor, right=of n4] {};
              
              \node (rej_n1) [dot, fill=lightBlueColor, below=of n1] {};
              \node (rej_n2) [dot, fill=lightBlueColor, below=of n2] {};
              \node (rej_n3) [dot, fill=lightBlueColor, below=of n3] {};
              \node (rej_n4) [dot, fill=lightBlueColor, below=of n4] {};
              
              \draw[-Latex, black] (root) to node [below, black] {start} (n1);
              \draw[-Latex, black] (n1) to (n2);
              \draw[-Latex, black] (n2) to (n3);
              \draw[-Latex, black] (n3) to (n4);
              
              \draw[-Latex, rejColor] (n1) to node [pos=0.5, right, rejColor] {rej} (rej_n1);
              \draw[-Latex, rejColor] (n2) to node [pos=0.5, right, rejColor] {rej} (rej_n2);
              \draw[-Latex, rejColor] (n3) to node [pos=0.5, right, rejColor] {rej} (rej_n3);
              \draw[-Latex, rejColor] (n4) to node [pos=0.5, right, rejColor] {rej} (rej_n4);
              
              \draw[-Latex, accColor, bend left=30] (n1) to node [pos=0.5, above, accColor] {acc} (n2);
              \draw[-Latex, accColor, bend left=30] (n2) to node [pos=0.5, above, accColor] {acc} (n3);
              \draw[-Latex, accColor, bend left=30] (n3) to node [pos=0.5, above, accColor] {acc} (n4);
              \draw[-Latex, accColor, bend left=30] (n4) to node [pos=0.5, above, accColor] {acc} (n5);
            \end{tikzpicture}
        }
    \end{minipage}\hfill 
    \begin{minipage}{0.2\textwidth} 
        \centering
        \subcaptionbox{\batch ($k=4$)\label{fig:batch}}{
        \begin{tikzpicture}[
          node distance=0.8cm and 1cm,
          dot/.style={circle,inner sep=2pt, label={center:}}
        ]
          
          \node (root) [dot, fill=black, opacity=0.3] at (0,0) {};

          \node (n1)   [dot, fill=black,  right=of root] {};
          \node (n2)   [dot, fill=black, below=of n1] {};
          \node (n3)   [dot, fill=black, below=of n2] {};
          \node (n4)   [dot, fill=black, below=of n3] {};
          \node (n5)   [dot, fill=lightBlueColor, below=of n4] {};

          \node (n1b)  [dot, fill=lightBlueColor, right=of n1] {};
          \node (n2b)  [dot, fill=lightBlueColor, right=of n2] {};
          \node (n3b)  [dot, fill=lightBlueColor, right=of n3] {};
          \node (n4b)  [dot, fill=lightBlueColor, right=of n4] {};

          \draw[-Latex, black] (root) to node [above, black] {start} (n1);
          \draw[-Latex, black] (root) to (n2);
          \draw[-Latex, black] (root) to (n3);
          \draw[-Latex, black] (root) to (n4);

          \draw[-Latex, rejColor] (n1) to node [pos=0.5, right, yshift=4pt, rejColor] {rej} (n2);
          \draw[-Latex, rejColor] (n2) to node [pos=0.5, right, yshift=4pt, rejColor] {rej} (n3);
          \draw[-Latex, rejColor] (n3) to node [pos=0.5, right, yshift=4pt, rejColor] {rej} (n4);
          \draw[-Latex, rejColor] (n4) to node [pos=0.5, right, yshift=4pt, rejColor] {rej} (n5);

          \draw[-Latex, accColor] (n1) to node [pos=0.5, above, accColor] {acc} (n1b);
          \draw[-Latex, accColor] (n2) to node [pos=0.5, above, accColor] {acc} (n2b);
          \draw[-Latex, accColor] (n3) to node [pos=0.5, above, accColor] {acc} (n3b);
          \draw[-Latex, accColor] (n4) to node [pos=0.5, above, accColor] {acc} (n4b);
        \end{tikzpicture}
        }
    \end{minipage}\hfill 
    \begin{minipage}{0.4\textwidth} 
        \centering
        \subcaptionbox{\tree ($k=6$)\label{fig:tree}}{
        \begin{tikzpicture}[
          node distance=0.8cm and 1cm,
          dot/.style={circle,inner sep=2pt, label={center:}}
        ]
          \node (r)   [dot, fill=gray, opacity=0.3] at (0,0) {};
          \node (A0)  [dot, fill=black, right=of r] {};
          \node (A3)  [dot, fill=black, below=of A0, yshift=-2.4cm] {};
          \node (B0)  [dot, fill=black, right=of A0] {};
          \node (B1)  [dot, fill=black, below=of B0, yshift=-0.8cm] {};
          \node (B3)  [dot, fill=black, right=of A3] {};
          \node (D0)  [dot, fill=black, right=of B0] {};

          \node (rej_A3)  [dot, fill=lightBlueColor, below=of A3] {};
          \node (rej_B3)  [dot, fill=lightBlueColor, below=of B3] {};
          \node (rej_B1)  [dot, fill=lightBlueColor, below=of B1] {};
          \node (rej_D0)  [dot, fill=lightBlueColor, below=of D0] {};

          \node (acc_D0)  [dot, fill=lightBlueColor, right=of D0] {};
          \node (acc_B1)  [dot, fill=lightBlueColor, right=of B1] {};
          \node (acc_B3)  [dot, fill=lightBlueColor, right=of B3] {};

          \draw[-Latex, black] (r)  to node [above, black] {start} (A0);
          \draw[-Latex, black] (r)  to (A3);
          \draw[-Latex, black] (A0) to (B0);
          \draw[-Latex, black] (A0) to (B1);
          \draw[-Latex, black] (A3) to (B3);
          \draw[-Latex, black] (B0) to (D0);

          \draw[-Latex, rejColor] (A0) to node [pos=0.5, right, rejColor] {rej} (A3);
          \draw[-Latex, rejColor] (B0) to node [pos=0.5, right, rejColor] {rej} (B1);
          \draw[-Latex, rejColor] (D0) to node [pos=0.5, right, rejColor] {rej} (rej_D0);
          \draw[-Latex, rejColor] (B1) to node [pos=0.5, right, rejColor] {rej} (rej_B1);
          \draw[-Latex, rejColor] (B3) to node [pos=0.5, right, rejColor] {rej} (rej_B3);
          \draw[-Latex, rejColor] (A3) to node [pos=0.5, right, rejColor] {rej} (rej_A3);

          \draw[-Latex, accColor, bend left=30] (A0) to node [pos=0.5, above, accColor] {acc} (B0);
          \draw[-Latex, accColor, bend left=30] (A3) to node [pos=0.5, above, accColor] {acc} (B3);
          \draw[-Latex, accColor, bend left=30] (B0) to node [pos=0.5, above, accColor] {acc} (D0);
          \draw[-Latex, accColor, bend left=30] (D0) to node [pos=0.5, above, accColor] {acc} (acc_D0);
          \draw[-Latex, accColor, bend left=30] (B1) to node [pos=0.5, above, accColor] {acc} (acc_B1);
          \draw[-Latex, accColor, bend left=30] (B3) to node [pos=0.5, above, accColor] {acc} (acc_B3);
        \end{tikzpicture}
        }
    \end{minipage} 
    \caption{Speculative decoding trees: a black draft tree overlaid with a green/red \acs{gsd} decision tree. Black vertices and arrows represent the draft tree; each vertex is a drafted token, and paths from the gray root are potential text continuations. Green/red arrows show \acs{gsd} acceptance/rejection decisions. Blue leaf vertices signify sampling from a distribution. (Note: \acs{ersd} does not follow the same decision tree.)}
    \label{fig:all_strategies}
\end{figure*}
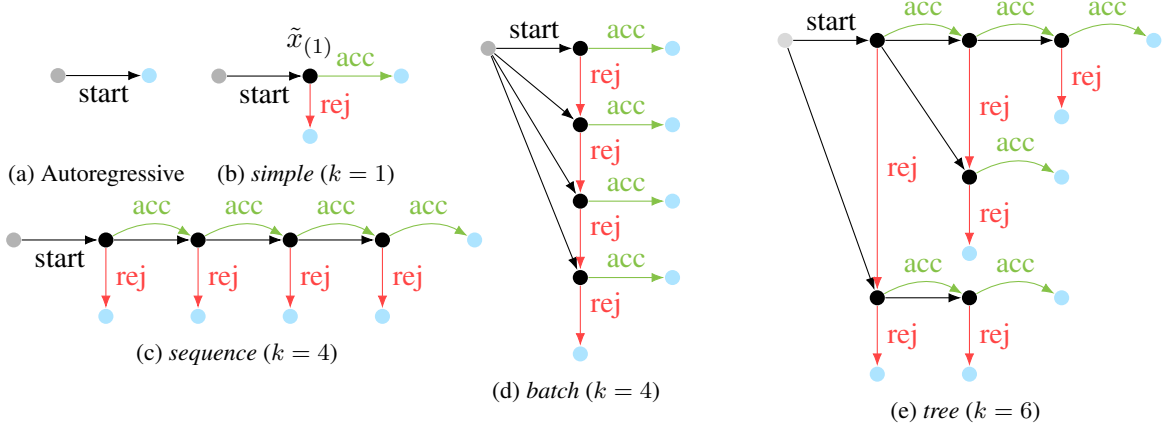
\section{Introduction}\label{s:Intro}
Transformer-based \gls{llm}s are at the forefront of the AI revolution, driving rapid advancements across numerous applications.  However, as their adoption accelerates, the speed of text generation emerges as a critical bottleneck, alongside compute and memory constraints.
Standard \gls{llm} generation involves calculating the conditional probability distribution using the target model $P$ given a partial output $x_{:n} = (x_1, x_2, ..., x_n)$, which is the concatenation of the initial context and all previously generated tokens. 
The target model $P$ outputs the conditional probability distribution $P(\:\cdot\mid x_{:n})$, from which a single token $x_{n+1}$ is sampled and concatenated to $x_{:n}$ to form the updated sequence $x_{:n+1}$. These steps are repeated iteratively until a stopping condition is met, such as reaching a maximum sequence length or generating an end-of-sequence token. This autoregressive token-by-token approach inherently leads to slow generation speeds.
Speculative decoding accelerates this process by potentially generating more than one token for each target model $P$ evaluation, while guaranteeing the same output quality--that is, the distribution of generated text remains identical to standard autoregressive decoding.
It consists of the following three steps:
\begin{enumerate}
    \item \textbf{Drafting}--a smaller draft model $Q$ is used to generate one or more possible continuations of the token sequence $x_{:n}$; this may involve multiple sampling calls of the draft model.
    \item \textbf{Evaluation} -- the target model $P$ evaluates all drafted continuations in parallel.
    \item \textbf{Verification} -- one or more tokens are accepted, based on their probabilities under $Q$ and $P$.
\end{enumerate}
Speculative decoding achieves a $\mathbf{2}$-$\mathbf{3}\times$ speed-up in text generation \citep{leviathan_fast_2023, chen_accelerating_2023}. Following common computational models, we quantify speed-up as the expected number of generated tokens per evaluation of the target model $P$. For a fair comparison of different speculative decoding strategies, we fix the number of drafted tokens $k$ and analyze the speed-up achieved.
Channel simulation is a compression problem focused on efficiently generating samples from a target probability distribution $P$ at a decoder \citep{li_channel_2024}. While only the encoder knows the target distribution $P$, both the encoder and decoder can access to samples from common reference distribution $Q$. Unlike standard source coding, the goal is not to communicate a specific sample, but rather to ensure that the decoder can generate any sample distributed according to $P$. Common methods rely on generating a shared codebook (list) of samples at the encoder and decoder. Based on the target distribution $P$, the encoder communicates to the decoder an index from this list, guaranteed to follow the target distribution $P$. A good introduction is provided by \citet{theis_algorithms_2022}, and \citet{li_channel_2024} compiles a comprehensive literature review.
Intuitively, both speculative decoding and channel simulation aim to generate samples from some target distribution $P$ by generating samples from a draft/reference distribution $Q$. In channel simulation, the goal is to minimize the entropy of the chosen index. In speculative decoding, the goal is to maximize the probability of one of the drafted tokens being accepted. Although these objectives seem different, the most common speculative decoding method \citep{leviathan_fast_2023, chen_accelerating_2023}, which we refer to as \gls{gsd}, (explained in the next section as a \batch drafting strategy), and a channel simulation method called greedy rejection sampling \citep{harsha_communication_2010, flamich_adaptive_2023} \textit{are essentially the same procedures!} To explain this surprising connection, we develop a theory that explicitly links the entropy of the accepted token with the speed-up of speculative decoding by considering the generation of multiple tokens. Additionally, we propose a novel speculative decoding strategy, \gls{ersd}, based on another channel simulation method called Poisson functional representation \citep{li_strong_2018}. We demonstrate that its performance matches that of \gls{gsd}. 
Our contributions in this paper are:
\begin{itemize}
    \item Establish a surprising connection between speculative decoding and channel simulation.
    \item Propose a new speculative decoding method using Poisson functional representation.
    \item Derive an explicit relation between the number of drafted tokens and the expected number of accepted tokens, i.e., the speed-up of response generation.
\end{itemize}
Throughout this paper, $\Omega$ denotes the finite set of tokens, where each token is represented by an integer corresponding to a textual element. Sets are enclosed in braces $\{\}$, sequences in parentheses $(\,)$ and concatenation of sequences $a$ and $b$ is written as $a \cat b$.
\section{Speculative decoding} \label{sec:speculative_decoding}
\begin{figure*}[t]
  \begin{minipage}[t]{0.49\linewidth}
    \begin{algorithm}[H]
      \caption{Simple \gls{gsd} ($k=1$)}
      \label{alg:greedy_speculative_decoding_simple}
      \begin{algorithmic}[1]
        \State \textbf{Input:} partial output $x_{:n}$, draft model $Q$, target model $P$ 
        \State \Call{Evaluate}{$Q, x_{:n}$} \label{line:greedy_sd_simple_draft_beg}
        \State $\xd \sim Q(\:\cdot\mid x_{:n})$ \label{line:greedy_sd_simple_draft_end}
        
        \State \Call{Evaluate}{$P, x_{:n} \cat (\xd)$} \label{line:greedy_sd_simple_evaluate}
        
        \State $U\sim \text{Uniform}(0, 1)$ \label{line:greedy_sd_simple_verfication_beg}
        \If{$\frac{P(\xd\mid x_{:n})}{Q(\xd\mid x_{:n})} > U$}
          \State $x_2^{next} \sim P(\:\cdot\mid x_{:n} \cat (\xd)))$
          \State \textbf{Return} $(\xd, x_2^{next})$
        \EndIf
        \State $P_{residual} \gets \max{(P(\:\cdot\mid x_{:n}) - Q(\:\cdot\mid x_{:n}), 0})$
        \State $P_{residual} \gets P_{residual} / \Call{sum}{P_{residual}}$
        \State $x_{residual} \sim P_{residual}$
        \State \textbf{Return} $(x_{residual})$ \label{line:greedy_sd_simple_verfication_end}
      \end{algorithmic}
    \end{algorithm}
  \end{minipage}\hfill
  \begin{minipage}[t]{0.49\linewidth}
    \begin{algorithm}[H]
      \caption{Simple \gls{ersd} ($k=1$)}
      \label{alg:exp_race_speculative_decoding_simple}
      \begin{algorithmic}[1]
        \State \textbf{Input:} partial output $x_{:n}$, draft model $Q$, target model $P$
        \State \Call{Evaluate}{$Q, x_{:n}$}
        \ForAll{$i \in \Omega$}
          \State $e_i \sim \text{Exp}(1)$
        \EndFor
        \State $\xd \gets \argmin_{i\in\Omega} \frac{e_i}{Q(i \mid x_{:n})}$
        
        \State \Call{Evaluate}{$P, x_{:n} \cat (\xd)$}
        
        \State $x^{next}_1 \gets \argmin_{i\in\Omega} \frac{e_i}{P(i \mid x_{:n})}$
        \If{$\xd = x^{next}_1$}
          \State $x^{next}_2 \sim P(\:\cdot\mid x_{:n}\cat (\xd))$
          \State \textbf{Return} $(\xd, x^{next}_2)$
        \EndIf
        \State \textbf{Return} $(x^{next}_1)$
      \end{algorithmic}
    \end{algorithm}
  \end{minipage}
\end{figure*}
\gls{llm}s predict the distributions of tokens conditioned on the preceding sequence. We can consider an LLM as a black box that takes a sequence of tokens $x_{:n}$ as input, and outputs a set of distributions $\{P(\:\cdot\mid x_{:i})\}_{i\leq n}$. In standard autoregressive decoding, we typically utilize only the last distribution $P(\:\cdot\mid x_{:n})$ to sample the next token. In contrast, speculative decoding leverages distributions conditioned on different subsequences to accelerate generation.
The simplest case of \gls{gsd}, is outlined in Algorithm \ref{alg:greedy_speculative_decoding_simple}. In the \textbf{drafting} stage (lines \ref{line:greedy_sd_simple_draft_beg}-\ref{line:greedy_sd_simple_draft_end}), a single token $\xd$ is generated using the draft model $Q$, we dub this \simple drafting strategy. Subsequently, in the \textbf{evaluation} stage (line \ref{line:greedy_sd_simple_evaluate}), the target model processes the sequence $x_{:n}\cat(\xd)$ and produces, among others, two relevant distributions: $P(\:\cdot\mid x_{:n})$ and $P(\:\cdot\mid x_{:n}\cat(\xd))$. The \textbf{verification} stage then determines whether to accept $\xd$ based on probabilities of both $P(\:\cdot\mid x_{:n})$ and $Q(\:\cdot\mid x_{:n})$. If $\xd$ is accepted, the algorithm effectively generates two tokens per target model evaluation, as $P(\:\cdot\mid x_{:n}\cat(\xd))$ is already computed and can be sampled from. Otherwise, if $\xd$ is rejected, the algorithm performs a standard autoregressive step, but samples from a \textit{residual distribution} $P_{residual}$ derived from $P(\:\cdot\mid x_{:n})$ and $Q(\:\cdot\mid x_{:n})$. This residual distribution focuses on sampling from tokens to which the target model $P$ assigns proportionally higher probability than the draft model $Q$.
A straightforward extension to the \simple drafting strategy is to speculate a \sequence of $k$ tokens instead of just one. During the \textbf{drafting} stage the draft model $Q$ is run autoregressively $k$ times to generate a draft sequence of length $k, (\xd_{(1)}, \xd_{(1,1)}, \ldots , \xd_{(1)_{i=1}^k})$. The notation $\xd_{\idx}$ describes that the token at position $\idx=(j_1, \dots, j_{m-1}, j_m)$ comes after $\xd_{\idx^\prime}$ where $\idx^\prime=(j_1, \dots, j_{m-1})$. The \textbf{evaluation} step then processes the entire drafted sequence $x_{:n}\cat (\xd_{(1)}, \xd_{(2)}, ..., \xd_{(k)})$ in parallel using the target model $P$. The drafted tokens are then \textbf{verified} and accepted sequentially, starting from the first, until a rejection occurs or all $k$ drafted tokens are accepted. Consequently, the \sequence strategy can generate from $1$ to $k+1$ tokens per evaluation of the target model $P$.
An alternative to \sequence drafting is \batch drafting \citep{sun2024block}, where multiple candidate continuations are explored. Specifically, $Q$ is evaluated once to sample multiple draft token alternatives, which are then evaluated in parallel by the target model $P$. While naively computationally intensive, \citet{miao_specinfer_2024} showed that it can be done at the same computational cost as \sequence drafting. \footnote{A straightforward implementation of \batch drafting would require evaluation of $k$ sequences of length $(n+1)$, but this can be reduced to a single sequence of length $(n+k)$ by manipulating attention masks and token embeddings -- resulting in the same cost as \sequence drafting.} This yields distributions $\left(P(\:\cdot\mid x_{:n} \cat (\xd_{(i)}))\right)_{i\leq k}$. The verification stage then sequentially considers each drafted token $\xd_{(i)}$ until the first acceptance. This strategy generates two tokens if any drafted token is accepted, or one if all get rejected. To avoid redundancy, candidates should be sampled without replacement \citep{jeon2024recursive}. To recall the connection highlighted in Section \ref{s:Intro}, the acceptance criterion in greedy rejection sampling and \batch \gls{gsd} when $k=|\Omega|$ is exactly the same. 
The \tree drafting strategy, which subsumes \simplex, \sequencex, and \batch methods, operates on any ordered tree topology. Each vertex represents a token drawn from $Q$. Each path from the root to a leaf constitutes a possible sequence drafted autoregressively. The general \tree \gls{gsd} algorithm, detailed in Appendix \ref{appendix:algo} and Algorithm \ref{alg:greedy_speculative_decoding_tree}, recursively applies \batch selection at each vertex of the tree to generate the accepted sequence. All the strategies are illustrated in Figure \ref{fig:all_strategies}.
The work most closely resembling ours is \citet{chen_sequoia_2025}, which investigates optimal tree topologies for \gls{gsd} under specific assumptions. Our analysis is more general and encompasses their problem setting as a special case.  Moreover, by establishing connections to information theory -- specifically channel simulation \citep{li_channel_2024} and Tunstall coding \citep{tunstall1968synthesis} -- we derive an upper bound on the expected speed-up, which converges when the number of drafted tokens $k$ is large. Additionally, our algorithm for generating decoding trees achieves $O(k \log k)$ complexity, outperforming the $O(k^2 |\Omega|)$ complexity of \citet{chen_sequoia_2025}. Additionally, we consider token-by-token generation, unlike global generation \citep{hu2024accelerated} which operates on trees natively.
\section{Exponential Races}\label{s:ExpRaces}
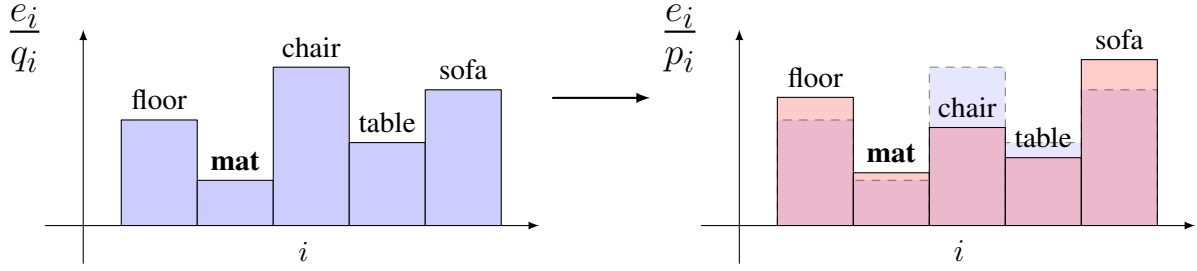
\begin{figure*}[t]
Context: \textbf{The cat sat on the} \vspace{1em}

\begin{tikzpicture}[
    >=latex,
    every node/.style={inner sep=1pt, outer sep=0pt, minimum size=0pt},
    every path/.style={line cap=round, line join=round},
  ]
  \draw[fill=blue!20] (0.5,0) -- (0.5,1.4) -- (1.5,1.4) -- (1.5,0);
  \draw[fill=blue!20] (1.5,0) -- (1.5,0.6) -- (2.5,0.6) -- (2.5,0);
  \draw[fill=blue!20] (2.5,0) -- (2.5,2.1) -- (3.5,2.1) -- (3.5,0);
  \draw[fill=blue!20] (3.5,0) -- (3.5,1.1) -- (4.5,1.1) -- (4.5,0);
  \draw[fill=blue!20] (4.5,0) -- (4.5,1.8) -- (5.5,1.8) -- (5.5,0);
  \draw[->] (-0.5,0) -- (6,0) node[right] {};
  \draw[->] (-0,-0.5) -- (-0,2.6) node[above] {};
  \node[anchor=south east] at (-0.5,2) {\huge $\frac{e_i}{q_i}$};
  \node[anchor=south east] at (3,-0.5) {$i$};
  \node[anchor=south] at (1, 1.5) {floor};
  \node[anchor=south] at (2, 0.7) {\textbf{mat}};
  \node[anchor=south] at (3, 2.2) {chair};
  \node[anchor=south] at (4, 1.2) {table};
  \node[anchor=south] at (5, 1.9) {sofa};
\end{tikzpicture}
\begin{tikzpicture}[
    >=latex,
    every node/.style={inner sep=1pt, outer sep=0pt, minimum size=0pt},
    every path/.style={line cap=round, line join=round},
    remember picture,
  ]
   \draw[->, thick] (0,0) -- (0.08\textwidth, 0);
   \node[inner sep=0pt, outer sep=0pt, minimum size=0pt] (node_name) at (0,0.2) {};
   \node[inner sep=0pt, outer sep=0pt, minimum size=0pt] (node_name) at (0,-2.2) {};
\end{tikzpicture}
\begin{tikzpicture}[
    >=latex,
    every node/.style={inner sep=1pt, outer sep=0pt, minimum size=0pt},
    every path/.style={line cap=round, line join=round},
  ]
  \begin{scope}[draw=gray!100]
  \draw[fill=blue!10, dashed] (0.5,0) - - (0.5,1.4) -- (1.5,1.4) -- (1.5,0);
  \draw[fill=blue!10, dashed] (1.5,0) -- (1.5,0.6) -- (2.5,0.6) -- (2.5,0) ;
  \draw[fill=blue!10, dashed] (2.5,0) -- (2.5,2.1) -- (3.5,2.1) -- (3.5,0);
  \draw[fill=blue!10, dashed] (3.5,0) -- (3.5,1.1) -- (4.5,1.1) -- (4.5,0);
  \draw[fill=blue!10, dashed] (4.5,0) -- (4.5,1.8) -- (5.5,1.8) -- (5.5,0);
  \end{scope}
  \draw[fill=red!100, fill opacity=0.2] (0.5,0) -- (0.5,1.7) -- (1.5,1.7) -- (1.5,0);
  \draw[fill=red!100, fill opacity=0.2] (1.5,0) -- (1.5,0.7) -- (2.5,0.7) -- (2.5,0);
  \draw[fill=red!100, fill opacity=0.2] (2.5,0) -- (2.5,1.3) -- (3.5,1.3) -- (3.5,0);
  \draw[fill=red!100, fill opacity=0.2] (3.5,0) -- (3.5,0.9) -- (4.5,0.9) -- (4.5,0);
  \draw[fill=red!100, fill opacity=0.2] (4.5,0) -- (4.5,2.2) -- (5.5,2.2) -- (5.5,0);
  \node[anchor=south] at (1, 1.8) {floor};
  \node[anchor=south] at (2, 0.8) {\textbf{mat}};
  \node[anchor=south] at (3, 1.4) {chair};
  \node[anchor=south] at (4, 1.0) {table};
  \node[anchor=south] at (5, 2.3) {sofa};
  \draw[->] (-0.5,0) -- (6,0) node[right] {};
  \draw[->] (-0,-0.5) -- (-0,2.6) node[above] {};
  \node[anchor=south east] at (3,-0.5) {\large $i$};
  \node[anchor=south east] at (-0.5,2) {\huge $\frac{e_i}{p_i}$};
\end{tikzpicture}
\caption {Illustration of exponential races for speculative decoding. Each bar represents a potential next token, with height corresponding to arrival time. The first arrival under the draft model distribution $Q$ (left) predicts the first arrival under the target model distribution $P$ (right).}
\label{fig:exp_race}
\end{figure*}
An exponential race \citep{maddison_poisson_2017} is a Poisson process with time-ordered points, each corresponding to a sample from a distribution $P$. We are interested in the winner (first point) of this race. In the discrete case, we can simulate relevant points, i.e., potential winners, by associating each element $i$ of the sample space $\Omega$ with an exponential random variable $e_i\sim \text{Exp}(1)$, $\mathbf{e}=\{e_i\}_{i\in\Omega}$. The winner of the race, $i^*=\argmin_{i\in\Omega} \frac{e_i}{p_i}$, is distributed according to $i^*\sim P$. This is also known as the Gumbel-max trick \citep{jang2017categorical}, where arrival times are obtained via the monotonic transformation $-\log(\cdot)$. Thus, given a distribution $P$, exponential races allow us to sample from it using independent exponential random variables.
Let $P$ and $Q$ be distributions with the same support $\Omega$. Using the same realization of $\mathbf{e}$ for exponential races yields $i^*_P=\argmin_{i\in\Omega} \frac{e_i}{p_i} \sim P$ and $i^*_Q=\argmin_{i\in\Omega} \frac{e_i}{q_i} \sim Q$, both following their respective distributions. If $P$ and $Q$ are similar (i.e., the ratio $\frac{p_i}{q_i}$ is close to $1$), then it is likely that $i^*_P=i^*_Q$. An example is illustrated in Figure \ref{fig:exp_race}, where $Q$ and $P$ are distributions generated by \gls{llm}s.
The simplest version of \gls{ersd} is presented in Algorithm \ref{alg:exp_race_speculative_decoding_simple}, where only a single token is generated from the draft model in the \textbf{drafting} stage: a winner of an exponential race under $Q(\;\cdot\mid x_{:n})$. As in \gls{gsd}, the target model $P$ is then \textbf{evaluated} on the sequence $x_{:n} \cat (\xd)$. During \textbf{verification}, if the winner of the exponential race under $P(\;\cdot\mid x_{:n})$ is also $\xd$, one more token is drafted.
It is straightforward to extend this approach to a \sequence strategy, by generating an exponential race for every node in the draft sequence. For the \batch strategy with $k$ alternatives, the drafted tokens are selected as the first $k$ arrivals of the exponential race under the draft model $Q$, as those are the most likely winners of the race under law $P$.
Like the traditional \gls{gsd}, the \batch case can be generalized to the full \tree by continuing different trajectories. The general procedure for this \tree approach is presented in Appendix \ref{appendix:algo} as Algorithm \ref{alg:exp_race_speculative_decoding_tree}.
Standard \gls{gsd} prioritizes maximizing the probability of accepting the initially proposed draft token.  Indeed, for the \simple (and thus, \sequence) drafting strategy, greedy decoding achieves optimality within token-based speculation \citep{sun_spectr_2023}, with the acceptance probability given by $1-D_{TV}[P(\,\cdot\mid x_{:n}), Q(\,\cdot\mid x_{:n})]$, where $D_{TV}[P,Q]=\frac{1}{2}\sum_{i\in\Omega}|P(i)-Q(i)|$ is the total variation distance \citep{csiszar2011information}. We can establish a corresponding bound for the acceptance probability of the first drafted token in \gls{ersd} using the Poisson matching lemma \citep{li_unified_2021}.
\begin{lemma} \label{lemma:exp_race_first_arrival}
    For \gls{ersd} with target distribution $P=P(\,\cdot\mid x_{:n})$ and draft distribution $Q=Q(\,\cdot\mid x_{:n})$, the probability of accepting the first drafted token, $P_{accept}^{(1)}$, satisfies:
    \begin{align}
        1-D_{TV}[P,Q]\geq P_{accept}^{(1)}\geq D_{HM}[P,Q],
    \end{align}
    where $D_{HM}$ denotes the harmonic mean distance, defined as:
    \begin{align}
        D_{HM}[P,Q] \stackrel{\text{def}}{=} \sum_{i\in\Omega} \frac{P(i)Q(i)}{P(i)+Q(i)}.
    \end{align}
\end{lemma}
The proof Lemma \ref{lemma:exp_race_first_arrival} is shown in Appendix \ref{appendix:proof_exp_race}. Despite a lower first-token acceptance probability, exponential races are not inferior to greedy decoding. As we will show, the overall performance for both methods hinges on the entropy of the acceptance distribution. From a channel simulation perspective, this entropy is linked to the Kullback-Leibler (KL) divergence between $P$ and $Q$.
\section{Markov Chain Simulation}
The question we are trying to answer in this section is: what is the best drafting strategy, i.e., what drafting tree is optimal? In such a tree, each vertex, except the root, is associated with a drafted token. Specifically, for a vertex indexed by $\idx = (j_1, \dots, j_l)$, the associated token $\xd_{\idx}$ is sampled from $Q(\:\cdot \mid x_{:n} \cat (\xd_{\idx_{:1}}, \dots, \xd_{\idx_{:l-1}}) )$, where $\idx_{:m} = (j_1, \dots, j_m)$. Let $R(\,\idx \mid x_{:n})$ denote the probability that vertex $\idx$, and thus the drafted token $\xd_{\idx}$, is accepted during verification, given the context $x_{:n}$ and a chosen speculative decoding algorithm. $R(\,\idx \mid x_{:n})$ is defined as the probability of acceptance, marginalized over all possible draft token sequences from $Q$:
\begin{align}
R(\,\idx \mid &x_{:n}) = \E_{\xd \sim Q(\cdot\mid x_{:n})}  
 \pr{\xd_\textbf{j} \text{ accepted} \mid x_{:n}, \xd}  
\end{align}
For the rest of the section, we assume the target and draft model distributions, $P$ and $Q$, be $m$-th order Markov sources.
Then, the acceptance probability $R(\,\idx \mid x_{:n})$ is an $m$-th order Markov chain, since $P$ and $Q$ are as well.
The expected number of accepted tokens is the sum of acceptance probabilities for each vertex in the drafting tree. Therefore, the optimal drafting tree $\tau^*$ with $k+1$ vertices (for speculating $k$ tokens) is the solution to:
\begin{align}
    \tau^*=\argmax_{\text{tree } \tau, |\tau|=k+1} \sum_{\idx \in \tau} R(\,\idx \mid x_{:n}).
\end{align}
Crucially, for any vertex $\idx$ and its descendant $\idx^\prime$ in a drafting tree, the acceptance probability is decreasing: $R(\,\idx \mid x_{:n})\geq R(\,\idx^\prime \mid x_{:n})$. This is because $\idx^\prime$ can only be accepted if its ancestor $\idx$ is accepted. Consequently, the top $k$ vertices with the highest acceptance probabilities form a valid tree. Any node $\idx$ in this top-$k$ list will have all its ancestors $\idx_{:m}$ (where $m<|\idx|$) also present in the list due to their higher acceptance probabilities. This observation leads to a greedy algorithm (Algorithm \ref{alg:tree}) for constructing an optimal tree with $k+1$ vertices. Algorithm \ref{alg:tree} iteratively builds the tree by greedily adding the next most likely token to be accepted. This is achieved efficiently using a priority queue to maintain candidate vertices, ordered by their acceptance probabilities. The algorithm's computational complexity is $O(k \log k)$ because the loop iterates $k$ times, and each iteration involves priority queue operations (push and pop) with a maximum queue size of $2k$, each taking $O(\log k)$ time.
\begin{algorithm}[t]
  \caption{Optimal tree construction}
  \label{alg:tree}
  \begin{algorithmic}[1]
    \State \textbf{Input:} \# drafted tokens $k$, partial output $x_{:n}$
    \State $tree \gets \{()\}$ \Comment{initialize tree with root only}
    \State $C \gets \Call{PriorityQueue}{\,}$ 
    
    \State \Call{Add}{$C, \left(R((0)\mid x_{:n}), (0)\right)$}
    \ForAll{$i \in (1,\dots,k)$}
        \State $\_, \idx \gets \Call{PopMax}{C}$
        \State $tree \gets tree \cup \{\idx\}$

        \State $\idx_{child} \gets \idx \cat (0)$
        \State $C.$\Call{Add}{$\left(R(\idx_{child}\mid x_{:n}), \idx_{child}\right)$}
        \If{$\idx_{|\idx|} < |\Omega|$}
            \State $\idx_{sibling} \gets \idx_{:(|\idx|-1)} \cat (\idx_{|\idx|}+1)$
            \State $C.$\Call{Add}{$\left(R(\idx_{sibling}\mid x_{:n}), \idx_{sibling}\right)$}
        \EndIf
    \EndFor
    \State \textbf{Return} tree
  \end{algorithmic}
\end{algorithm}
Our analysis relies on the acceptance probability function $R$, which is typically unknown in practice. If we approximate $R$ using an empirical acceptance distribution -- effectively treating it as a 0-th order Markov source -- the resulting algorithm becomes equivalent to that proposed by \citet{chen_sequoia_2025}. We employ this empirical approximation in our experiments. However, as Section \ref{sec:experiments} will demonstrate, this 0-th order approximation proves to be inaccurate, leading to an underestimation of the speed-up we can achieve with speculative decoding. Developing more refined approximations of $R$ presents a promising avenue for future investigation.
In another connection to information theory, the process of generating optimal drafting trees (Algorithm \ref{alg:tree}) closely resembles the construction of Tunstall codes \citep{tunstall1968synthesis}. Due to their similarity, quantities in speculative decoding, such as the number of drafted tokens, speed-up, and entropy of the acceptance distribution, have direct counterparts in Tunstall codes: the number of expanded nodes in the Tunstall tree, the expected length of the consumed source symbols, and the source entropy, respectively. This connection allows us to express the expected speed-up of speculative decoding (both \gls{gsd} and \gls{ersd}) in terms of these quantities.
Tunstall codes are a type of variable-to-fixed length source codes used in data compression for discrete sources. In source coding, the goal is to represent sequences of symbols from a source alphabet (like tokens in our case) using the fewest number of bits (in the case of a binary alphabet). Variable-to-fixed length codes, such as Tunstall codes, achieve this by mapping variable-length sequences of source symbols to fixed-length codewords from a code alphabet. Specifically, a Tunstall code takes a variable-length prefix of the source symbol sequence and encodes it into a fixed-length output sequence. This encoding step is repeated until the entire source sequence is processed. Tunstall codes are known to be optimal in the sense that for sufficiently long codewords, the average number of output alphabet symbols per source symbol approaches the entropy of the source -- this also holds true for sources with memory \citep{savari_generalized_1997}.
\begin{figure*}[t]
  \centering
  \includegraphics[width=0.95\textwidth]{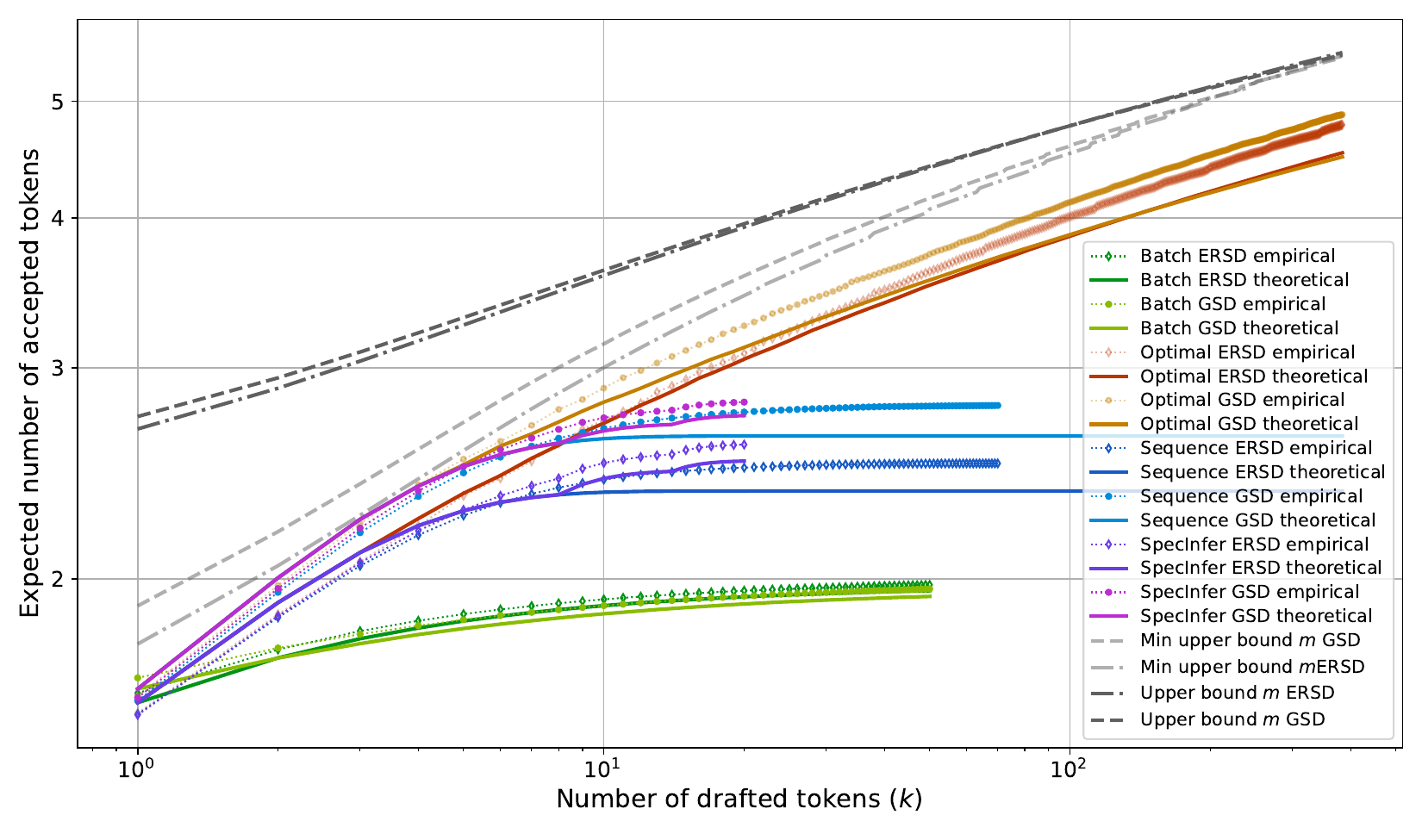}
  \vspace{-0.5em}
  \caption{Expected number of accepted tokens as a function of the number of drafted tokens for \sequence, \batch, $\tau^*$ \tree (optimal), and SpecInfer \tree drafting strategies, for \gls{gsd} and \gls{ersd}. Results shown for draft model $Q$ Llama-3.2-1B, and target model $P$ Llama-3.1-70B-Instruct.}
  \label{fig:expected-accepted-tokens}
\end{figure*}
\begin{figure*}[t]
  \centering
  \includegraphics[width=0.95\textwidth]{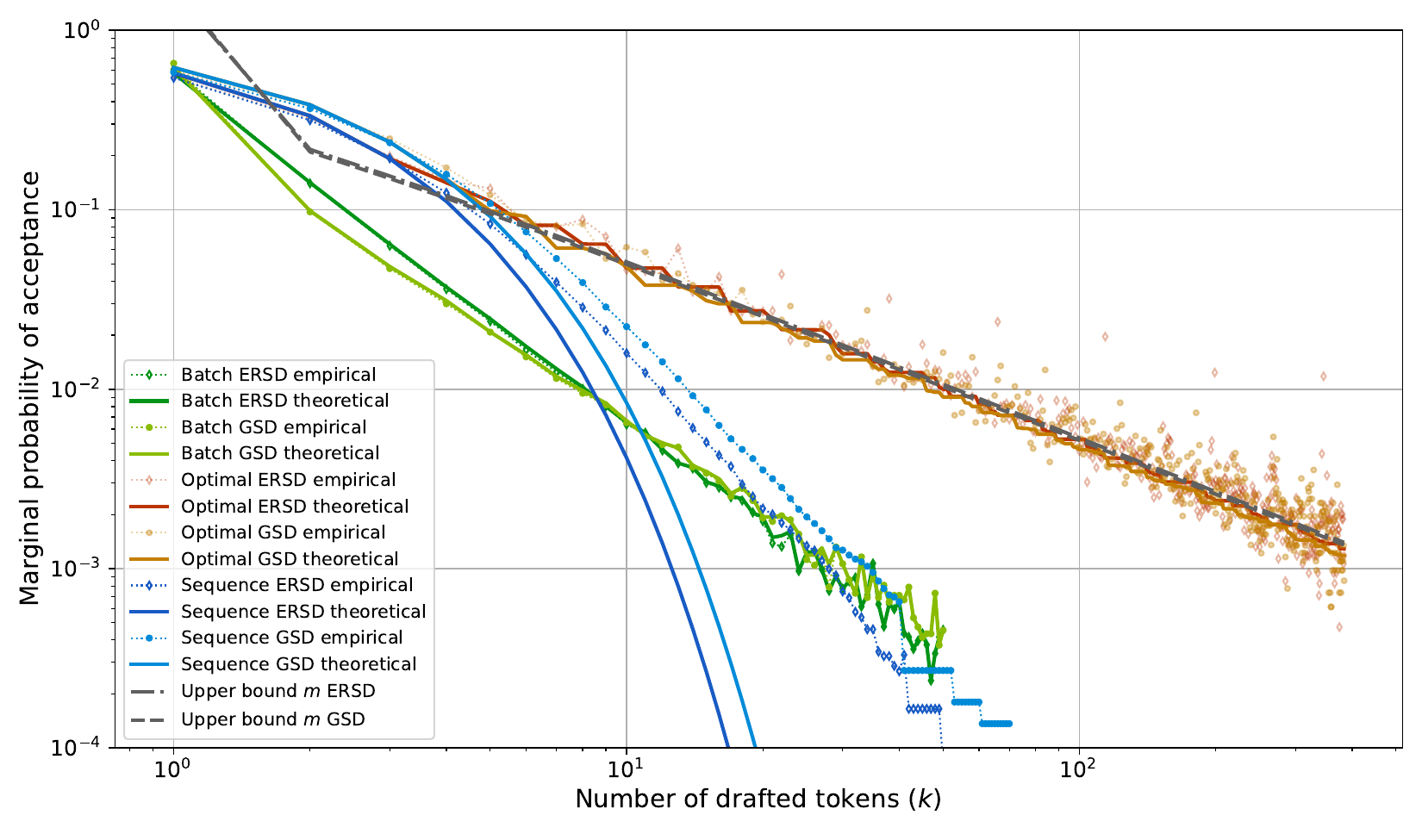}
  \vspace{-0.5em}
  \caption{Marginal probability of acceptance as a function of the number of drafted tokens for \sequence, \batch, $\tau^*$ \tree (optimal) drafting strategies, for \gls{gsd} and \gls{ersd}. Results shown for draft model $Q$ Llama-3.2-1B, and target model $P$ Llama-3.1-70B-Instruct.}
  \label{fig:marginal-acceptance-prob}
\end{figure*}
Let the entropy of the acceptance probability distribution be defined as:
\begin{align}
    \entb{R}=\E_{x_{:n}\sim P} \entb{R(\,\cdot\mid x_{:n})}
\end{align}
\begin{theorem}{(Upper bound Tunstall)} \label{thm:upperbound}
    For speculative decoding employing the optimal drafting strategy $\tau^*$, the expected number of generated tokens is bounded as follows:
    \begin{align}
        \E\left[\text{\# generated tokens}\right] \leq \frac{\log|\Omega| + \log (k+1)}{\entb{R}}\,. \nonumber
    \end{align}

\end{theorem}
Details of Theorem \ref{thm:upperbound} are provided in Appendix \ref{appendix:upperbound_tunstall}. As the theorem suggests, for a large number of drafted tokens $k$, the generation speed-up is fundamentally governed by the entropy of the acceptance distribution, $\entb{R}$. Channel simulation algorithms, such as greedy rejection sampling and Poisson functional representation, are explicitly designed to minimize this entropy. Consequently, these methods, which underpin both \gls{gsd} and \gls{ersd}, inherently aim to maximize generation speed-up by reducing $\entb{R}$. From a channel simulation perspective, the KL divergence $\D\left[P\|Q\right]$ between the target distribution $P$ and the draft distribution $Q$ provides a lower bound on this crucial entropy \citep{li_channel_2024}:
\begin{align}
    \D\left[P\|Q\right] \leq \entb{R}.
\end{align}
Furthermore, channel simulation theory provides upper bounds on $\entb{R}$ in terms of $\D\left[P\|Q\right]$ for greedy rejection coding \citep{harsha_communication_2010} (relevant to \gls{gsd}):
\begin{align}
    \entb{R} \leq& \D\left[P\|Q\right] + \\ &(1+\epsilon) \log(\D\left[P\|Q\right]+1) + O(1), \nonumber
\end{align}
for any $\epsilon>0$, and for Poisson functional representation \cite{li_strong_2018} (relevant to \gls{ersd}):
\begin{align}
    \entb{R} \leq& \D\left[P\|Q\right] + \\ &\log(\D\left[P\|Q\right]+1) + 4\,. \nonumber
\end{align}
These bounds show the connection of speed-up in speculative decoding to KL divergence $\D\left[P\|Q\right]$ between the models.
While asymptotically accurate for large $k$, the bound from Theorem \ref{thm:upperbound} is dominated by the full token alphabet size $|\Omega|$ when the number of drafted tokens $k$ is small. For such values of $k$, we observe that the optimal drafting tree typically explores only a limited number of drafting positions. Indeed, multiple acceptance distributions can yield identical optimal drafting trees and the same acceptance probabilities for all nodes within those trees. Thus, evaluating the bound from Theorem \ref{thm:upperbound} for any such distribution provides a valid upper bound for all of them. To simplify the notation and exposition, we focus on the 0-th order acceptance distribution $R$, noting that our findings can be generalized to higher-order Markov sources.
For a given number of drafted tokens $k$, let $d$ be the maximum index explored in the optimal drafting tree $\tau^*_k$ with $k+1$ vertices, i.e., $d = \max_{\idx\in\tau^*_k} \max_{i} \{i \mid i \in \idx \}$. We define an equivalent acceptance distribution $\hat{R}$ such that for indices $i \leq d$, $\hat{R}(i) = R(i)$, and for $i > d$, $\hat{R}(i) \leq \min_{i^\prime \leq d} R(i^\prime)$. In short, $\hat{R}$ matches $R$ on tree nodes for the first $d$ indices and is upper-bounded by $\min_{i^\prime \leq d} R(i^\prime)$ thereafter.
Our objective is to minimize the upper bound on the expected number of accepted tokens. Upon inspection of Theorem \ref{thm:upperbound}, we see that a tighter bound is achieved by maximizing the entropy of $\hat{R}$ and minimizing the alphabet size. Minimizing the bound requires strategically distributing the remaining probability mass, $p_{res} = 1 - \sum_{i=1}^d R(i)$, associated with indices beyond $d$. For any choice to incorporate $m$ additional indices, entropy maximization is achieved by distributing the residual mass equally across them.
\begin{lemma}{(Upper bound $m$)} \label{lemma:upperbound_m}
    For a fixed number of drafted tokens $k$ and initial index range $d$, consider a 0-th order acceptance probability distribution $R$. Let $p_{res} = 1 - \sum_{i=1}^d R(i)$ be the residual probability mass. Define  $R^{res}$ such that $R^{res}(i) = R(i)$ for $i \leq d$, and $R^{res}(d+1) = p_{res}$.
    Then, for any integer $m \geq \frac{p_{res}}{\min_{i \leq d} R(i)}$, the expected number of generated tokens is upper-bounded by:
    \begin{align}
        \E\left[\text{\# generated tokens}\right] \leq \frac{\log(d+m) + \log (k+1)}{\entb{R^{res}} + p_{res} \log m}. \nonumber
    \end{align}
\end{lemma}
\vspace{-0.5em}
Note: $R^{res}$ defined in above lemma is not $\hat{R}$.
\section{Experiments} \label{sec:experiments}
To validate our theoretical analysis, we conducted numerical experiments comparing \gls{gsd} and \gls{ersd} across different drafting strategies. We performed open-ended text generation up to $200$ tokens, accumulating $100k$ generated tokens per strategy. The experiments were performed on 8 Nvidia RTX A6000 GPUs, with experiments running for $140$ hours of wall-clock time.
We estimated the acceptance probability function $R$ using empirical acceptance probabilities for different indices (from \sequence and \batch), approximating it as a $0$-th order Markov chain. Based on this estimated $R$, we computed the optimal drafting tree $\tau^*$ for both \gls{gsd} and \gls{ersd} using Algorithm \ref{alg:tree}.
Figure \ref{fig:expected-accepted-tokens} illustrates the expected number of accepted tokens as a function of drafted tokens for the target model Llama-3.1 70B and the draft model Llama-3.2 1B \citep{grattafiori_llama_2024}, used under the Meta Llama 3.1 and 3.2 Licenses. Figure \ref{fig:marginal-acceptance-prob} shows the marginal change in acceptance probability with each additional drafted token. We compare four drafting strategies: \batch, \sequence, optimal \tree ($\tau^*$), and the SpecInfer \tree \citep{miao_specinfer_2024}, which drafts 3-sequences with the first two drafted tokens being common.
For each strategy and speculative decoding method (\gls{gsd}, \gls{ersd}), we present both theoretical and empirical performance. The theoretical plots show the expected number of generated tokens based on the estimated $R$, while the empirical plots display the actual value observed in our experiments.
Figure \ref{fig:expected-accepted-tokens} reveals that for \batch drafting with $k=1$, \gls{gsd} outperforms \gls{ersd}, consistent with Lemma~\ref{lemma:exp_race_first_arrival}. However, with $k=2$, both methods exhibit comparable performance. For $k\geq3$, \gls{ersd} achieves a higher expected number of accepted tokens. Both methods plateau just below $2$, as anticipated. The acceptance probabilities for each index ($(0),(1),(2),\dots$) are shown in Figure \ref{fig:marginal-acceptance-prob}. For \sequence drafting, \gls{gsd} demonstrates superior performance compared to \gls{ersd}, owing to \sequence's focus on the first arrivals. Intriguingly, our empirical results also challenge the 0-th order acceptance assumption \citep{chen_sequoia_2025}.  The empirical performance plateaus at a higher level than predicted by the measured $R$, indicating higher-order Markov dependencies.  This can be explained by regions of language where the target model $P$ and draft model $Q$ exhibit greater alignment, leading to extended sequences of accepted tokens.  The SpecInfer \tree achieves slightly improved performance over \sequence drafting but still plateaus.
As predicted, the optimal $\tau^*$ \tree strategy yields the best performance for both \gls{gsd} and \gls{ersd}, exhibiting a logarithmic relationship between drafted tokens $k$ and expected generated tokens (Figure \ref{fig:expected-accepted-tokens}). At low $k$, the optimal \tree $\tau^*$ and \sequence strategies show similar performance, with \gls{gsd} slightly outperforming \gls{ersd}. However, as $k$ increases, the performance gap diminishes, and they converge. 
Figure \ref{fig:expected-accepted-tokens} presents the minimal upper bound derived from Lemma~\ref{lemma:upperbound_m} by optimizing the parameter $m$ for each number of drafted tokens $k$. We also depict the upper bound from Lemma~\ref{lemma:upperbound_m}, calculated with a fixed $m$ for $k=384$ to show the behavior of the bound. While the marginal changes in the minimal upper bound are omitted for visual clarity--due to their step-like transitions--, we observe that the marginal changes of the fixed-$m$ upper bound closely follow the trends of the optimal $\tau^*$ \tree. 
\section{Limitations}
Speculative decoding's speed gains are most significant when drafting only a few tokens, $k$.  Therefore, for small values $k$, \gls{gsd} is often the most practical choice. While token-by-token generation was the focus of this work, joint sequence generation in speculative decoding or channel simulation could lead to further improvements, but its computational practicality is uncertain. 
Furthermore, our current stateless approximation of the acceptance probability function $R$ oversimplifies the contextual nature of language. Developing more context-aware approximations could yield improvements in future work.
\section{Conclusion}
This work establishes a connection between speculative decoding – a technique for accelerating autoregressive \gls{llm} generation – and channel simulation.  This connection enabled us to propose \gls{ersd}, a novel speculative decoding method. By linking the optimal drafting strategy to Tunstall codes, we derived a theoretical upper bound and the asymptotic relationship between the number of speculated tokens and the expected speed-up, for \gls{gsd} and \gls{ersd}. These findings offer a deeper understanding of, and potential improvements to, the efficiency of speculative decoding.
\bibliography{cite.bib}

\begin{thebibliography}{20}
\providecommand{\natexlab}[1]{#1}

\bibitem[{Chen et~al.(2023)Chen, Borgeaud, Irving, Lespiau, Sifre, and Jumper}]{chen_accelerating_2023}
Charlie Chen, Sebastian Borgeaud, Geoffrey Irving, Jean-Baptiste Lespiau, Laurent Sifre, and John Jumper. 2023.
\newblock \href {https://doi.org/10.48550/arXiv.2302.01318} {Accelerating {Large} {Language} {Model} {Decoding} with {Speculative} {Sampling}}.
\newblock \emph{arXiv preprint}.
\newblock ArXiv:2302.01318 [cs].

\bibitem[{Chen et~al.(2025)Chen, May, Svirschevski, Huang, Ryabinin, Jia, and Chen}]{chen_sequoia_2025}
Zhuoming Chen, Avner May, Ruslan Svirschevski, Yu-Hsun Huang, Max Ryabinin, Zhihao Jia, and Beidi Chen. 2025.
\newblock \href {https://proceedings.neurips.cc/paper_files/paper/2024/hash/ea1f5f0878d43ff4fb8bf64ef4a2326c-Abstract-Conference.html} {Sequoia: {Scalable} and {Robust} {Speculative} {Decoding}}.
\newblock \emph{Advances in Neural Information Processing Systems}, 37:129531--129563.

\bibitem[{Csisz{\'a}r and K{\"o}rner(2011)}]{csiszar2011information}
Imre Csisz{\'a}r and J{\'a}nos K{\"o}rner. 2011.
\newblock \emph{Information Theory: Coding Theorems for Discrete Memoryless Systems}, 2 edition.
\newblock Cambridge University Press.

\bibitem[{Flamich and Theis(2023)}]{flamich_adaptive_2023}
Gergely Flamich and Lucas Theis. 2023.
\newblock \href {https://doi.org/10.1109/ISIT54713.2023.10206725} {Adaptive {Greedy} {Rejection} {Sampling}}.
\newblock In \emph{2023 {IEEE} {International} {Symposium} on {Information} {Theory} ({ISIT})}, pages 454--459.
\newblock ISSN: 2157-8117.

\bibitem[{Grattafiori et~al.(2024)Grattafiori, Dubey, Jauhri, Pandey, Kadian, Al-Dahle, Letman, Mathur, Schelten, Vaughan, Yang, Fan, Goyal, Hartshorn, Yang, Mitra, Sravankumar, Korenev, Hinsvark, Rao, Zhang, Rodriguez, Gregerson, Spataru, Roziere, Biron, Tang, Chern, Caucheteux, Nayak, Bi, Marra, McConnell, Keller, Touret, Wu, Wong, Ferrer, Nikolaidis, Allonsius, Song, Pintz, Livshits, Wyatt, Esiobu, Choudhary, Mahajan, Garcia-Olano, Perino, Hupkes, Lakomkin, AlBadawy, Lobanova, Dinan, Smith, Radenovic, Guzmán, Zhang, Synnaeve, Lee, Anderson, Thattai, Nail, Mialon, Pang, Cucurell, Nguyen, Korevaar, Xu, Touvron, Zarov, Ibarra, Kloumann, Misra, Evtimov, Zhang, Copet, Lee, Geffert, Vranes, Park, Mahadeokar, Shah, Linde, Billock, Hong, Lee, Fu, Chi, Huang, Liu, Wang, Yu, Bitton, Spisak, Park, Rocca, Johnstun, Saxe, Jia, Alwala, Prasad, Upasani, Plawiak, Li, Heafield, Stone, El-Arini, Iyer, Malik, Chiu, Bhalla, Lakhotia, Rantala-Yeary, Maaten, Chen, Tan, Jenkins, Martin, Madaan, Malo, Blecher, Landzaat,
  Oliveira, Muzzi, Pasupuleti, Singh, Paluri, Kardas, Tsimpoukelli, Oldham, Rita, Pavlova, Kambadur, Lewis, Si, Singh, Hassan, Goyal, Torabi, Bashlykov, Bogoychev, Chatterji, Zhang, Duchenne, Çelebi, Alrassy, Zhang, Li, Vasic, Weng, Bhargava, Dubal, Krishnan, Koura, Xu, He, Dong, Srinivasan, Ganapathy, Calderer, Cabral, Stojnic, Raileanu, Maheswari, Girdhar, Patel, Sauvestre, Polidoro, Sumbaly, Taylor, Silva, Hou, Wang, Hosseini, Chennabasappa, Singh, Bell, Kim, Edunov, Nie, Narang, Raparthy, Shen, Wan, Bhosale, Zhang, Vandenhende, Batra, Whitman, Sootla, Collot, Gururangan, Borodinsky, Herman, Fowler, Sheasha, Georgiou, Scialom, Speckbacher, Mihaylov, Xiao, Karn, Goswami, Gupta, Ramanathan, Kerkez, Gonguet, Do, Vogeti, Albiero, Petrovic, Chu, Xiong, Fu, Meers, Martinet, Wang, Wang, Tan, Xia, Xie, Jia, Wang, Goldschlag, Gaur, Babaei, Wen, Song, Zhang, Li, Mao, Coudert, Yan, Chen, Papakipos, Singh, Srivastava, Jain, Kelsey, Shajnfeld, Gangidi, Victoria, Goldstand, Menon, Sharma, Boesenberg, Baevski,
  Feinstein, Kallet, Sangani, Teo, Yunus, Lupu, Alvarado, Caples, Gu, Ho, Poulton, Ryan, Ramchandani, Dong, Franco, Goyal, Saraf, Chowdhury, Gabriel, Bharambe, Eisenman, Yazdan, James, Maurer, Leonhardi, Huang, Loyd, Paola, Paranjape, Liu, Wu, Ni, Hancock, Wasti, Spence, Stojkovic, Gamido, Montalvo, Parker, Burton, Mejia, Liu, Wang, Kim, Zhou, Hu, Chu, Cai, Tindal, Feichtenhofer, Gao, Civin, Beaty, Kreymer, Li, Adkins, Xu, Testuggine, David, Parikh, Liskovich, Foss, Wang, Le, Holland, Dowling, Jamil, Montgomery, Presani, Hahn, Wood, Le, Brinkman, Arcaute, Dunbar, Smothers, Sun, Kreuk, Tian, Kokkinos, Ozgenel, Caggioni, Kanayet, Seide, Florez, Schwarz, Badeer, Swee, Halpern, Herman, Sizov, Guangyi, Zhang, Lakshminarayanan, Inan, Shojanazeri, Zou, Wang, Zha, Habeeb, Rudolph, Suk, Aspegren, Goldman, Zhan, Damlaj, Molybog, Tufanov, Leontiadis, Veliche, Gat, Weissman, Geboski, Kohli, Lam, Asher, Gaya, Marcus, Tang, Chan, Zhen, Reizenstein, Teboul, Zhong, Jin, Yang, Cummings, Carvill, Shepard, McPhie, Torres,
  Ginsburg, Wang, Wu, U, Saxena, Khandelwal, Zand, Matosich, Veeraraghavan, Michelena, Li, Jagadeesh, Huang, Chawla, Huang, Chen, Garg, A, Silva, Bell, Zhang, Guo, Yu, Moshkovich, Wehrstedt, Khabsa, Avalani, Bhatt, Mankus, Hasson, Lennie, Reso, Groshev, Naumov, Lathi, Keneally, Liu, Seltzer, Valko, Restrepo, Patel, Vyatskov, Samvelyan, Clark, Macey, Wang, Hermoso, Metanat, Rastegari, Bansal, Santhanam, Parks, White, Bawa, Singhal, Egebo, Usunier, Mehta, Laptev, Dong, Cheng, Chernoguz, Hart, Salpekar, Kalinli, Kent, Parekh, Saab, Balaji, Rittner, Bontrager, Roux, Dollar, Zvyagina, Ratanchandani, Yuvraj, Liang, Alao, Rodriguez, Ayub, Murthy, Nayani, Mitra, Parthasarathy, Li, Hogan, Battey, Wang, Howes, Rinott, Mehta, Siby, Bondu, Datta, Chugh, Hunt, Dhillon, Sidorov, Pan, Mahajan, Verma, Yamamoto, Ramaswamy, Lindsay, Lindsay, Feng, Lin, Zha, Patil, Shankar, Zhang, Zhang, Wang, Agarwal, Sajuyigbe, Chintala, Max, Chen, Kehoe, Satterfield, Govindaprasad, Gupta, Deng, Cho, Virk, Subramanian, Choudhury, Goldman,
  Remez, Glaser, Best, Koehler, Robinson, Li, Zhang, Matthews, Chou, Shaked, Vontimitta, Ajayi, Montanez, Mohan, Kumar, Mangla, Ionescu, Poenaru, Mihailescu, Ivanov, Li, Wang, Jiang, Bouaziz, Constable, Tang, Wu, Wang, Wu, Gao, Kleinman, Chen, Hu, Jia, Qi, Li, Zhang, Zhang, Adi, Nam, Yu, Wang, Zhao, Hao, Qian, Li, He, Rait, DeVito, Rosnbrick, Wen, Yang, Zhao, and Ma}]{grattafiori_llama_2024}
Aaron Grattafiori, Abhimanyu Dubey, Abhinav Jauhri, Abhinav Pandey, Abhishek Kadian, Ahmad Al-Dahle, Aiesha Letman, Akhil Mathur, Alan Schelten, Alex Vaughan, Amy Yang, Angela Fan, Anirudh Goyal, Anthony Hartshorn, Aobo Yang, Archi Mitra, Archie Sravankumar, Artem Korenev, Arthur Hinsvark, Arun Rao, Aston Zhang, Aurelien Rodriguez, Austen Gregerson, Ava Spataru, Baptiste Roziere, Bethany Biron, Binh Tang, Bobbie Chern, Charlotte Caucheteux, Chaya Nayak, Chloe Bi, Chris Marra, Chris McConnell, Christian Keller, Christophe Touret, Chunyang Wu, Corinne Wong, Cristian~Canton Ferrer, Cyrus Nikolaidis, Damien Allonsius, Daniel Song, Danielle Pintz, Danny Livshits, Danny Wyatt, David Esiobu, Dhruv Choudhary, Dhruv Mahajan, Diego Garcia-Olano, Diego Perino, Dieuwke Hupkes, Egor Lakomkin, Ehab AlBadawy, Elina Lobanova, Emily Dinan, Eric~Michael Smith, Filip Radenovic, Francisco Guzmán, Frank Zhang, Gabriel Synnaeve, Gabrielle Lee, Georgia~Lewis Anderson, Govind Thattai, Graeme Nail, Gregoire Mialon, Guan Pang,
  Guillem Cucurell, Hailey Nguyen, Hannah Korevaar, Hu~Xu, Hugo Touvron, Iliyan Zarov, Imanol~Arrieta Ibarra, Isabel Kloumann, Ishan Misra, Ivan Evtimov, Jack Zhang, Jade Copet, Jaewon Lee, Jan Geffert, Jana Vranes, Jason Park, Jay Mahadeokar, Jeet Shah, Jelmer van~der Linde, Jennifer Billock, Jenny Hong, Jenya Lee, Jeremy Fu, Jianfeng Chi, Jianyu Huang, Jiawen Liu, Jie Wang, Jiecao Yu, Joanna Bitton, Joe Spisak, Jongsoo Park, Joseph Rocca, Joshua Johnstun, Joshua Saxe, Junteng Jia, Kalyan~Vasuden Alwala, Karthik Prasad, Kartikeya Upasani, Kate Plawiak, Ke~Li, Kenneth Heafield, Kevin Stone, Khalid El-Arini, Krithika Iyer, Kshitiz Malik, Kuenley Chiu, Kunal Bhalla, Kushal Lakhotia, Lauren Rantala-Yeary, Laurens van~der Maaten, Lawrence Chen, Liang Tan, Liz Jenkins, Louis Martin, Lovish Madaan, Lubo Malo, Lukas Blecher, Lukas Landzaat, Luke~de Oliveira, Madeline Muzzi, Mahesh Pasupuleti, Mannat Singh, Manohar Paluri, Marcin Kardas, Maria Tsimpoukelli, Mathew Oldham, Mathieu Rita, Maya Pavlova, Melanie Kambadur,
  Mike Lewis, Min Si, Mitesh~Kumar Singh, Mona Hassan, Naman Goyal, Narjes Torabi, Nikolay Bashlykov, Nikolay Bogoychev, Niladri Chatterji, Ning Zhang, Olivier Duchenne, Onur Çelebi, Patrick Alrassy, Pengchuan Zhang, Pengwei Li, Petar Vasic, Peter Weng, Prajjwal Bhargava, Pratik Dubal, Praveen Krishnan, Punit~Singh Koura, Puxin Xu, Qing He, Qingxiao Dong, Ragavan Srinivasan, Raj Ganapathy, Ramon Calderer, Ricardo~Silveira Cabral, Robert Stojnic, Roberta Raileanu, Rohan Maheswari, Rohit Girdhar, Rohit Patel, Romain Sauvestre, Ronnie Polidoro, Roshan Sumbaly, Ross Taylor, Ruan Silva, Rui Hou, Rui Wang, Saghar Hosseini, Sahana Chennabasappa, Sanjay Singh, Sean Bell, Seohyun~Sonia Kim, Sergey Edunov, Shaoliang Nie, Sharan Narang, Sharath Raparthy, Sheng Shen, Shengye Wan, Shruti Bhosale, Shun Zhang, Simon Vandenhende, Soumya Batra, Spencer Whitman, Sten Sootla, Stephane Collot, Suchin Gururangan, Sydney Borodinsky, Tamar Herman, Tara Fowler, Tarek Sheasha, Thomas Georgiou, Thomas Scialom, Tobias Speckbacher,
  Todor Mihaylov, Tong Xiao, Ujjwal Karn, Vedanuj Goswami, Vibhor Gupta, Vignesh Ramanathan, Viktor Kerkez, Vincent Gonguet, Virginie Do, Vish Vogeti, Vítor Albiero, Vladan Petrovic, Weiwei Chu, Wenhan Xiong, Wenyin Fu, Whitney Meers, Xavier Martinet, Xiaodong Wang, Xiaofang Wang, Xiaoqing~Ellen Tan, Xide Xia, Xinfeng Xie, Xuchao Jia, Xuewei Wang, Yaelle Goldschlag, Yashesh Gaur, Yasmine Babaei, Yi~Wen, Yiwen Song, Yuchen Zhang, Yue Li, Yuning Mao, Zacharie~Delpierre Coudert, Zheng Yan, Zhengxing Chen, Zoe Papakipos, Aaditya Singh, Aayushi Srivastava, Abha Jain, Adam Kelsey, Adam Shajnfeld, Adithya Gangidi, Adolfo Victoria, Ahuva Goldstand, Ajay Menon, Ajay Sharma, Alex Boesenberg, Alexei Baevski, Allie Feinstein, Amanda Kallet, Amit Sangani, Amos Teo, Anam Yunus, Andrei Lupu, Andres Alvarado, Andrew Caples, Andrew Gu, Andrew Ho, Andrew Poulton, Andrew Ryan, Ankit Ramchandani, Annie Dong, Annie Franco, Anuj Goyal, Aparajita Saraf, Arkabandhu Chowdhury, Ashley Gabriel, Ashwin Bharambe, Assaf Eisenman, Azadeh
  Yazdan, Beau James, Ben Maurer, Benjamin Leonhardi, Bernie Huang, Beth Loyd, Beto~De Paola, Bhargavi Paranjape, Bing Liu, Bo~Wu, Boyu Ni, Braden Hancock, Bram Wasti, Brandon Spence, Brani Stojkovic, Brian Gamido, Britt Montalvo, Carl Parker, Carly Burton, Catalina Mejia, Ce~Liu, Changhan Wang, Changkyu Kim, Chao Zhou, Chester Hu, Ching-Hsiang Chu, Chris Cai, Chris Tindal, Christoph Feichtenhofer, Cynthia Gao, Damon Civin, Dana Beaty, Daniel Kreymer, Daniel Li, David Adkins, David Xu, Davide Testuggine, Delia David, Devi Parikh, Diana Liskovich, Didem Foss, Dingkang Wang, Duc Le, Dustin Holland, Edward Dowling, Eissa Jamil, Elaine Montgomery, Eleonora Presani, Emily Hahn, Emily Wood, Eric-Tuan Le, Erik Brinkman, Esteban Arcaute, Evan Dunbar, Evan Smothers, Fei Sun, Felix Kreuk, Feng Tian, Filippos Kokkinos, Firat Ozgenel, Francesco Caggioni, Frank Kanayet, Frank Seide, Gabriela~Medina Florez, Gabriella Schwarz, Gada Badeer, Georgia Swee, Gil Halpern, Grant Herman, Grigory Sizov, Guangyi, Zhang, Guna
  Lakshminarayanan, Hakan Inan, Hamid Shojanazeri, Han Zou, Hannah Wang, Hanwen Zha, Haroun Habeeb, Harrison Rudolph, Helen Suk, Henry Aspegren, Hunter Goldman, Hongyuan Zhan, Ibrahim Damlaj, Igor Molybog, Igor Tufanov, Ilias Leontiadis, Irina-Elena Veliche, Itai Gat, Jake Weissman, James Geboski, James Kohli, Janice Lam, Japhet Asher, Jean-Baptiste Gaya, Jeff Marcus, Jeff Tang, Jennifer Chan, Jenny Zhen, Jeremy Reizenstein, Jeremy Teboul, Jessica Zhong, Jian Jin, Jingyi Yang, Joe Cummings, Jon Carvill, Jon Shepard, Jonathan McPhie, Jonathan Torres, Josh Ginsburg, Junjie Wang, Kai Wu, Kam~Hou U, Karan Saxena, Kartikay Khandelwal, Katayoun Zand, Kathy Matosich, Kaushik Veeraraghavan, Kelly Michelena, Keqian Li, Kiran Jagadeesh, Kun Huang, Kunal Chawla, Kyle Huang, Lailin Chen, Lakshya Garg, Lavender A, Leandro Silva, Lee Bell, Lei Zhang, Liangpeng Guo, Licheng Yu, Liron Moshkovich, Luca Wehrstedt, Madian Khabsa, Manav Avalani, Manish Bhatt, Martynas Mankus, Matan Hasson, Matthew Lennie, Matthias Reso, Maxim
  Groshev, Maxim Naumov, Maya Lathi, Meghan Keneally, Miao Liu, Michael~L. Seltzer, Michal Valko, Michelle Restrepo, Mihir Patel, Mik Vyatskov, Mikayel Samvelyan, Mike Clark, Mike Macey, Mike Wang, Miquel~Jubert Hermoso, Mo~Metanat, Mohammad Rastegari, Munish Bansal, Nandhini Santhanam, Natascha Parks, Natasha White, Navyata Bawa, Nayan Singhal, Nick Egebo, Nicolas Usunier, Nikhil Mehta, Nikolay~Pavlovich Laptev, Ning Dong, Norman Cheng, Oleg Chernoguz, Olivia Hart, Omkar Salpekar, Ozlem Kalinli, Parkin Kent, Parth Parekh, Paul Saab, Pavan Balaji, Pedro Rittner, Philip Bontrager, Pierre Roux, Piotr Dollar, Polina Zvyagina, Prashant Ratanchandani, Pritish Yuvraj, Qian Liang, Rachad Alao, Rachel Rodriguez, Rafi Ayub, Raghotham Murthy, Raghu Nayani, Rahul Mitra, Rangaprabhu Parthasarathy, Raymond Li, Rebekkah Hogan, Robin Battey, Rocky Wang, Russ Howes, Ruty Rinott, Sachin Mehta, Sachin Siby, Sai~Jayesh Bondu, Samyak Datta, Sara Chugh, Sara Hunt, Sargun Dhillon, Sasha Sidorov, Satadru Pan, Saurabh Mahajan,
  Saurabh Verma, Seiji Yamamoto, Sharadh Ramaswamy, Shaun Lindsay, Shaun Lindsay, Sheng Feng, Shenghao Lin, Shengxin~Cindy Zha, Shishir Patil, Shiva Shankar, Shuqiang Zhang, Shuqiang Zhang, Sinong Wang, Sneha Agarwal, Soji Sajuyigbe, Soumith Chintala, Stephanie Max, Stephen Chen, Steve Kehoe, Steve Satterfield, Sudarshan Govindaprasad, Sumit Gupta, Summer Deng, Sungmin Cho, Sunny Virk, Suraj Subramanian, Sy~Choudhury, Sydney Goldman, Tal Remez, Tamar Glaser, Tamara Best, Thilo Koehler, Thomas Robinson, Tianhe Li, Tianjun Zhang, Tim Matthews, Timothy Chou, Tzook Shaked, Varun Vontimitta, Victoria Ajayi, Victoria Montanez, Vijai Mohan, Vinay~Satish Kumar, Vishal Mangla, Vlad Ionescu, Vlad Poenaru, Vlad~Tiberiu Mihailescu, Vladimir Ivanov, Wei Li, Wenchen Wang, Wenwen Jiang, Wes Bouaziz, Will Constable, Xiaocheng Tang, Xiaojian Wu, Xiaolan Wang, Xilun Wu, Xinbo Gao, Yaniv Kleinman, Yanjun Chen, Ye~Hu, Ye~Jia, Ye~Qi, Yenda Li, Yilin Zhang, Ying Zhang, Yossi Adi, Youngjin Nam, Yu, Wang, Yu~Zhao, Yuchen Hao, Yundi
  Qian, Yunlu Li, Yuzi He, Zach Rait, Zachary DeVito, Zef Rosnbrick, Zhaoduo Wen, Zhenyu Yang, Zhiwei Zhao, and Zhiyu Ma. 2024.
\newblock \href {https://doi.org/10.48550/arXiv.2407.21783} {The {Llama} 3 {Herd} of {Models}}.
\newblock \emph{arXiv preprint}.
\newblock ArXiv:2407.21783 [cs].

\bibitem[{Harsha et~al.(2010)Harsha, Jain, McAllester, and Radhakrishnan}]{harsha_communication_2010}
Prahladh Harsha, Rahul Jain, David McAllester, and Jaikumar Radhakrishnan. 2010.
\newblock \href {https://doi.org/10.1109/TIT.2009.2034824} {{The {Communication} {Complexity} of {Correlation}}}.
\newblock \emph{IEEE Transactions on Information Theory}, 56(1):438--449.

\bibitem[{Hu and Huang(2024)}]{hu2024accelerated}
Zhengmian Hu and Heng Huang. 2024.
\newblock \href {https://openreview.net/forum?id=stMhi1Sn2G} {Accelerated speculative sampling based on tree monte carlo}.
\newblock In \emph{Forty-first International Conference on Machine Learning}.

\bibitem[{Jang et~al.(2017)Jang, Gu, and Poole}]{jang2017categorical}
Eric Jang, Shixiang Gu, and Ben Poole. 2017.
\newblock \href {https://openreview.net/forum?id=rkE3y85ee} {Categorical reparameterization with gumbel-softmax}.
\newblock In \emph{International Conference on Learning Representations}.

\bibitem[{Jeon et~al.(2024)Jeon, Gagrani, Goel, Park, Lee, and Lott}]{jeon2024recursive}
Wonseok Jeon, Mukul Gagrani, Raghavv Goel, Junyoung Park, Mingu Lee, and Christopher Lott. 2024.
\newblock \href {https://openreview.net/forum?id=RdKYAHZPxg} {Recursive speculative decoding: Accelerating {LLM} inference via sampling without replacement}.
\newblock In \emph{ICLR 2024 Workshop on Large Language Model (LLM) Agents}.

\bibitem[{Leviathan et~al.(2023)Leviathan, Kalman, and Matias}]{leviathan_fast_2023}
Yaniv Leviathan, Matan Kalman, and Yossi Matias. 2023.
\newblock \href {https://proceedings.mlr.press/v202/leviathan23a.html} {Fast {Inference} from {Transformers} via {Speculative} {Decoding}}.
\newblock In \emph{Proceedings of the 40th {International} {Conference} on {Machine} {Learning}}, pages 19274--19286. PMLR.
\newblock ISSN: 2640-3498.

\bibitem[{Li(2024)}]{li_channel_2024}
Cheuk~Ting Li. 2024.
\newblock \href {https://doi.org/10.1561/0100000141} {Channel {Simulation}: {Theory} and {Applications} to {Lossy} {Compression} and {Differential} {Privacy}}.
\newblock \emph{Foundations and Trends® in Communications and Information Theory}, 21(6):847--1106.
\newblock Publisher: Now Publishers, Inc.

\bibitem[{Li and Anantharam(2021)}]{li_unified_2021}
Cheuk~Ting Li and Venkat Anantharam. 2021.
\newblock \href {https://doi.org/10.1109/TIT.2021.3058842} {A {Unified} {Framework} for {One}-{Shot} {Achievability} via the {Poisson} {Matching} {Lemma}}.
\newblock \emph{IEEE Transactions on Information Theory}, 67(5):2624--2651.

\bibitem[{Li and El~Gamal(2018)}]{li_strong_2018}
Cheuk~Ting Li and Abbas El~Gamal. 2018.
\newblock \href {https://doi.org/10.1109/TIT.2018.2865570} {{Strong {Functional} {Representation} {Lemma} and {Applications} to {Coding} {Theorems}}}.
\newblock \emph{IEEE Transactions on Information Theory}, 64(11):6967--6978.

\bibitem[{Maddison(2017)}]{maddison_poisson_2017}
Chris~J. Maddison. 2017.
\newblock \href {https://ieeexplore.ieee.org/document/8093857} {A {Poisson} {Process} {Model} for {Monte} {Carlo}}.
\newblock In \emph{Perturbations, {Optimization}, and {Statistics}}, pages 193--231. MIT Press.
\newblock Conference Name: Perturbations, Optimization, and Statistics.

\bibitem[{Miao et~al.(2024)Miao, Oliaro, Zhang, Cheng, Wang, Zhang, Wong, Zhu, Yang, Shi, Shi, Chen, Arfeen, Abhyankar, and Jia}]{miao_specinfer_2024}
Xupeng Miao, Gabriele Oliaro, Zhihao Zhang, Xinhao Cheng, Zeyu Wang, Zhengxin Zhang, Rae Ying~Yee Wong, Alan Zhu, Lijie Yang, Xiaoxiang Shi, Chunan Shi, Zhuoming Chen, Daiyaan Arfeen, Reyna Abhyankar, and Zhihao Jia. 2024.
\newblock \href {https://doi.org/10.1145/3620666.3651335} {{SpecInfer}: {Accelerating} {Large} {Language} {Model} {Serving} with {Tree}-based {Speculative} {Inference} and {Verification}}.
\newblock In \emph{Proceedings of the 29th {ACM} {International} {Conference} on {Architectural} {Support} for {Programming} {Languages} and {Operating} {Systems}, {Volume} 3}, volume~3 of \emph{{ASPLOS} '24}, pages 932--949, New York, NY, USA. Association for Computing Machinery.

\bibitem[{Savari and Gallager(1997)}]{savari_generalized_1997}
S.A. Savari and R.G. Gallager. 1997.
\newblock \href {https://doi.org/10.1109/18.556121} {Generalized {Tunstall} codes for sources with memory}.
\newblock \emph{IEEE Transactions on Information Theory}, 43(2):658--668.

\bibitem[{Sun et~al.(2024)Sun, Mendlovic, Leviathan, Aharoni, Beirami, Ro, and Suresh}]{sun2024block}
Ziteng Sun, Uri Mendlovic, Yaniv Leviathan, Asaf Aharoni, Ahmad Beirami, Jae~Hun Ro, and Ananda~Theertha Suresh. 2024.
\newblock Block verification accelerates speculative decoding.
\newblock In \emph{Workshop on Efficient Systems for Foundation Models II @ ICML2024}.

\bibitem[{Sun et~al.(2023)Sun, Suresh, Ro, Beirami, Jain, and Yu}]{sun_spectr_2023}
Ziteng Sun, Ananda~Theertha Suresh, Jae~Hun Ro, Ahmad Beirami, Himanshu Jain, and Felix Yu. 2023.
\newblock \href {https://proceedings.neurips.cc/paper_files/paper/2023/hash/6034a661584af6c28fd97a6f23e56c0a-Abstract-Conference.html} {{SpecTr}: {Fast} {Speculative} {Decoding} via {Optimal} {Transport}}.
\newblock \emph{Advances in Neural Information Processing Systems}, 36:30222--30242.

\bibitem[{Theis and Yosri(2022)}]{theis_algorithms_2022}
Lucas Theis and Noureldin Yosri. 2022.
\newblock \href {https://proceedings.mlr.press/v162/theis22a.html} {{Algorithms for the {Communication} of {Samples}}}.
\newblock In \emph{{Proceedings of the 39th {International} {Conference} on {Machine} {Learning}}}, pages 21308--21328. PMLR.

\bibitem[{Tunstall(1968)}]{tunstall1968synthesis}
Brian Tunstall. 1968.
\newblock \emph{Synthesis of Noiseless Compression Codes}.
\newblock Ph.d. thesis, Georgia Institute of Technology.

\end{thebibliography}
\onecolumn
\appendix
\section{Algorithms} \label{appendix:algo}
\begin{algorithm}[H]
  \caption{\Acl{gsd}}
  \label{alg:greedy_speculative_decoding_tree}
  \begin{algorithmic}[1]
    \State \textbf{Input:} partial output $x_{:n}$, draft model $Q$, target model $P$, \# draft tokens $k$, draft strategy $\tau
    $
    
    \ForAll{$\idx \in \tau$} \Comment{Assumes lexicographical ordering of indexes $\idx$}
        \If{$\idx_{|\idx|}=1$} \Comment{If $\idx$ is the first child evaluate $Q$}
            \State \Call{Evaluate}{$Q, x_{:n} \cat (\xd_{\idx_{:i}})_{i=1}^{|\idx|-1}$}
        \EndIf
        \State $Q_{draft}=Q(\,\cdot\mid x_{:n} \cat (\xd_{\idx_{:i}})_{i=1}^{|\idx|-1})$
        
        \State $\idx_{anc} = \idx_{:|\idx|-1}$
        \ForAll{$i \in \N^+, i<\idx_{|\idx|}$} \Comment{Ensure sampling without replacement}
            \State $Q_{draft}(\xd_{\idx_{anc}\cat(i)}) \gets 0$
        \EndFor
        \State $Q_{draft} \gets Q_{draft}/\Call{sum}{Q_{draft}}$
        \State $\xd_\idx \sim Q_{draft}$
    \EndFor
    
    \State $\Call{Evaluate}{P, x_{:n}, \{\xd_\idx\}_{\idx \in \tau}}$ \Comment{Evaluate all draft tokens in parallel}
    \State $acc \gets \text{True}$
    \State $y \gets (\;)$
    \State $\idx \gets (\;)$
    \While{$acc$} 
        \State $acc \gets \text{False}$
        \State $P_{target} \gets P(\:\cdot\mid x_{:n} \cat y)$
        \State $Q_{draft} \gets Q(\:\cdot\mid x_{:n} \cat y)$
        \State $i\gets 1$ \Comment{Current considered child}
        \While{$\idx\cat(i) \in \tau $} 
            \If{$\frac{P_{target}(\xd_\idx)}{Q_{draft}(\xd_\idx)} > \text{Uniform}(0, 1)$}
            \Comment{Accept token and continue}
                \State $y \gets y \cup \{\xd_\idx\}$
                \State $acc \gets \text{True}$
                \State \textbf{break}
            \EndIf
            \State $P_{target} \gets \max(P_{target}-Q_{draft}, 0)$
            \State $P_{target} \gets P_{target} / \Call{sum}{P_{target}}$ \Comment{On rejection calculate residual distribution}
            \State $Q_{draft}(token) \gets 0$
            \State $Q_{draft} \gets Q_{draft} / \Call{sum}{Q_{draft}}$ \Comment{Update target distribution, to reflect sampling w.o. replacement}
            \State $i \gets i+1$
        \EndWhile
       \ForAll{$token \in S(y, x_{draft})$}
        \EndFor
    \EndWhile
    \State $token \gets \Call{sample}{P_{target}}$ \Comment{Accept token from residual distribution}
    \State $y \gets y \cup \{token\}$
    \State \textbf{return} $y$
  \end{algorithmic}
\end{algorithm}
The draft token selection step--sampling without replacement--in the general \gls{gsd} Algorithm \ref{alg:greedy_speculative_decoding_tree} can be implemented using exponential races just as in Algorithm \ref{alg:exp_race_speculative_decoding_tree}, or equivalently the Gumbel-max trick.  Furthermore, the verification step in \gls{ersd} Algorithm \ref{alg:exp_race_speculative_decoding_tree} is a significanly simpler compared to \gls{gsd}.
\begin{algorithm}[H]
  \caption{Exponential Race Speculative Decoding}
  \label{alg:exp_race_speculative_decoding_tree}
  \begin{algorithmic}[1]
    \State \textbf{Input:} partial output $x_{:n}$, draft model $Q$, target model $P$, \# draft tokens $k$, draft strategy $\tau$
    \ForAll{$\idx \in \tau$} \Comment{Assumes lexicographical ordering of indexes $\idx$}
        \State $\idx_{anc} = \idx_{:|\idx|-1}$
        \If{$\idx_{|\idx|}=1$} \Comment{If $\idx$ is the first child evaluate $Q$ and generate the race}
            \ForAll{$i \in \Omega$}
                \State $e_{\idx_{anc}\cat(i)} \gets \text{Exp}(1)$
            \EndFor
            \State \Call{Evaluate}{$Q, x_{:n} \cat (\xd_{\idx_{:i}})_{i=1}^{|\idx|-1}$}
        \EndIf
        \State $\xd_\idx \gets \operatorname{\idx_{|\idx|}-th} \argmin_{i\in\Omega} 
        \frac{e_{\idx_{anc}\cat(i)}}{Q(i\mid x_{:n} \cat (\xd_{\idx_{:i}})_{i=1}^{|\idx|-1})}$  \Comment{Find $\idx_{|\idx|}$-th race arrival under $Q$}    
    \EndFor
    \State $\Call{Evaluate}{P, x_{:n}, \{\xd_\idx\}_{\idx \in \tau}}$ \Comment{Evaluate all draft tokens in parallel}
    \State $y \gets (\;)$
    \State $\idx \gets (\;)$
    \While{$true$}
        \State $x^{next} \gets \argmin_{i\in\Omega} 
        \frac{e_{\idx_{anc}\cat(i)}}{Q(i\mid x_{:n} \cat (\xd_{\idx_{:i}})_{i=1}^{|\idx|-1})}$
        \Comment{Winner of race under $P$}
        \State $y \gets y \cat (x^{next})$
        \If{$x^{next} \notin \{\xd_{\idx \cat (i)} \mid i \in \Omega, \idx \cat (i) \in \tau \}$}
            \State \textbf{break}
        \EndIf
    \EndWhile
    \State \textbf{return} $y$
  \end{algorithmic}
\end{algorithm}
\section{Speculative decoding via exponential races} \label{appendix:proof_exp_race}
We begin by recalling the Poisson matching lemma \citep{li_unified_2021}, adapted to our notation for discrete alphabets:
\begin{lemma}{(Poisson matching lemma)} \label{lemma:pml}
Let $P$ and $Q$ be two probability distributions over the alphabet $\Omega$. Let $E_i \sim \text{Exp}(1)$ for each symbol $i \in \Omega$ be independent exponential random variables. Define $I^*_P = \argmin_{i\in\Omega} \frac{E_i}{p_i}$ and $I^*_Q = \argmin_{i\in\Omega} \frac{E_i}{q_i}$. The probability that $I^*_P$ is different than $I^*_Q$, given $I^*_Q$, is bounded by:
\begin{align}
     \pr{I^*_P\neq I^*_Q \mid I^*_Q} \leq 1 - \left(1 + \frac{q_{I^*_Q}}{p_{I^*_Q}}\right)^{-1}.
\end{align}
\end{lemma}
In essence, this lemma bounds the probability that two races, driven by the same underlying exponential random variables but with different distributions $P$ and $Q$, will have different winners.
\proof{(Lemma \ref{lemma:exp_race_first_arrival})
To obtain the average probability of differing first arrivals, we marginalize the Poisson matching lemma over all possible values of $I^*_Q$. Let $P_{accept}^{(1)}$ be the probability that the first drafted token in \gls{ersd} is accepted. This occurs when the winner of the race under $P$ is the same as the winner under $Q$, i.e., $I^*_P = I^*_Q$. Thus:
\begin{align}
    P_{accept}^{(1)} &= \pr{I^*_P = I^*_Q} \\
    &= 1 - \pr{I^*_P \neq I^*_Q} \\
    &= 1 - \E_{I^*_Q} \left[ \pr{I^*_P \neq I^*_Q|I^*_Q} \right ] \\
    &\geq 1 - \E_{I^*_Q} \left[ 1 - \left(1 + \frac{q_{I^*_Q}}{p_{I^*_Q}}\right)^{-1} \right] \quad \text{(by Lemma \ref{lemma:pml})} \\
    &= \E_{I^*_Q} \left[ \left(1 + \frac{q_{I^*_Q}}{p_{I^*_Q}}\right)^{-1} \right]\\
    &= \sum_{i\in \Omega} q_i \frac{1}{1+\frac{q_i}{p_i}} \quad \text{(since } I^*_Q \sim Q \text{)}\\
    &= \sum_{i\in \Omega} \frac{p_i q_i}{p_i + q_i} \\
    &= D_{HM}[P,Q],
\end{align}
where the last step follows from the definition of the harmonic mean distance $D_{HM}[P,Q]$.
Furthermore, it is known that $P_{accept}^{(1)}$ is upper-bounded by $1-D_{TV}[P, Q]$ (cite), and in general those bounds do not coincide as:
\begin{align}
    1-D_{TV}[P,Q] &= 1-\frac{1}{2}\sum_{i\in\Omega} |p_i-q_i| \\
           &= \frac{1}{2}\sum_{i\in\Omega}p_i + \frac{1}{2}\sum_{i\in\Omega}q_i -\frac{1}{2}\sum_{i\in\Omega} |p_i-q_i| \\
           &= \frac{1}{2}\sum_{i\in\Omega} \frac{(p_i+q_i)^2 - |p_i-q_i|(p_i+q_i)}{p_i+q_i} \\
           &= \frac{1}{2}\sum_{i\in\Omega} \frac{p_i^2 + 2p_iq_i + q_i^2 - |p_i^2-q_i^2|}{p_i+q_i} \\
           &\geq \frac{1}{2}\sum_{i\in\Omega} \frac{p_i^2 + 2p_iq_i + q_i^2 - (p_i^2+q_i^2)}{p_i+q_i} \quad \text{(since } |x| \geq x \text{)}\\
           &=\sum_{i\in\Omega} \frac{p_i q_i}{p_i+q_i} \\
           &=D_{HM}[P, Q].
\end{align}
}
\section{Tunstall Codes} \label{appendix:upperbound_tunstall}
This section provides a concise overview of Tunstall coding. Source coding is a fundamental technique for representing sequences of source symbols, like text or tokens, as sequences of bits (or symbols from another alphabet). The goal is efficient representation for storage or transmission. Tunstall coding is a variable-to-fixed length source coding method. This means it parses the source symbol sequence into variable-length subsequences, and then maps each of these subsequences to a fixed-length codeword. Let $G$ denote the expected length of the encoded source subsequence, i.e., $G = \E \left[\text{length of encoded source subsequence} \right]$.
Tunstall codes are constructed using trees. The construction process begins with a root node. Assuming a source alphabet $\Omega$, the root is expanded to have $|\Omega|$ children. Subsequently, in each step, the leaf node representing the most probable source sequence is expanded by adding $|\Omega|$ children to it. This expansion process is repeated until a desired number of codewords is reached. The structure and construction of this Tunstall tree, specifically its inner nodes, are identical to the optimal draft tree employed in speculative decoding as described in Algorithm \ref{alg:tree}; conversly, the leafs of the Tunstall tree correspond to sampling of an additinal token once no more drafted tokens are considered. If the construction process expands $k$ nodes, the resulting Tunstall tree will have $|\Omega| + k(|\Omega|-1)$ leaves. Each path from the root to a leaf represents a variable-length sequence of source symbols. By assigning a unique codeword of fixed length $\left\lceil{\log\left(k(|\Omega|-1) + |\Omega|\right)}\right\rceil$ bits to each leaf node, we create the Tunstall code.
Considering the fundamental limit of compression given by the source entropy, we can establish an inequality for compressing a source sequence of length $L$ with a Tunstall code:
\begin{align}
    L \entb{R} \leq \frac{L}{G} \log\left(k(|\Omega|-1) + |\Omega|\right)
\end{align}
Here, the LHS represents the theoretical minimum number of bits to encode a sequence of length $L$, and the RHS represents the expected number of bits used by the Tunstall code. The term $\frac{L}{G}$ represents the expected number of codewords needed to encode a source sequence of length $L$, and $\log\left(k(|\Omega|-1) + |\Omega|\right)$ is the fixed length of each codeword. Due to the connection between Tunstall code construction and Algorithm \ref{alg:tree}, specifically the optimal drafting tree, the 'length of encoded source sequence' equals 'number of generated tokens' in speculative decoding. Rearranging the equation, we obtain:
\begin{align}
    \E\left[\text{\# of generated tokens}\right] \leq& \frac{\log\left(k(|\Omega|-1) + |\Omega|\right)}{\entb{R}}
    \leq \frac{\log|\Omega| + \log (k+1)}{\entb{R}}.
\end{align}
Tunstall codes are known to be asymptotically optimal, even for sources with memory \cite{savari_generalized_1997}, when a different code for each state is used. This asymptotic optimality means that the gap to the theoretical compression limit becomes constant as $k$ increases. Therefore, this upper bound also characterizes the asymptotic behavior of speculative decoding.
\end{document}